\documentclass[10pt,twocolumn,letterpaper]{article}

\usepackage[pagenumbers]{cvpr} 

\usepackage{graphicx}
\usepackage{amsmath}
\usepackage{amssymb}
\usepackage{booktabs}
\usepackage{pifont}
\usepackage{setspace}
\usepackage{algpseudocode}

\usepackage[table]{xcolor}
\usepackage[pagebackref,breaklinks,colorlinks]{hyperref}

\usepackage[capitalize]{cleveref}
\crefname{section}{Sec.}{Secs.}
\Crefname{section}{Section}{Sections}
\Crefname{table}{Table}{Tables}
\crefname{table}{Tab.}{Tabs.}

\usepackage{bbding}
\usepackage{soul}
\usepackage{bibunits}
\usepackage{bm}
\usepackage{array}
\usepackage{xcolor}
\usepackage{pdflscape}
\usepackage{makecell}
\usepackage{framed,multirow}
\usepackage[flushleft]{threeparttable}
\usepackage{amssymb}
\usepackage{pifont}
\usepackage{algorithm}
\usepackage{layouts}
\usepackage{listings}
\usepackage{pythonhighlight}
\makeatletter
\newcommand\footnoteref[1]{\protected@xdef\@thefnmark{\ref{#1}}\@footnotemark}
\makeatother

\newcolumntype{P}[1]{>{\centering\arraybackslash}p{#1}}
\newlength\savewidth

\definecolor{Highlight}{HTML}{39b54a}  

\definecolor{igray}{rgb}{0.00, 0.00, 0.00}
\definecolor{iorange}{RGB}{232,158,51}
\definecolor{igreen}{RGB}{0,159,118}

\def\cvprPaperID{1212} 
\def\confName{CVPR}
\def\confYear{2023}

\begin{document}


\title{Post-Training Quantization for 3D Medical Image Segmentation: \\ 
A Practical Study on Real Inference Engines}

\author{******\\[2mm]
{\small Project:~\href{https://github.com/ljwztc/CLIP-Driven-Universal-Model}{Post-training quantization for 3D medical image Segmentation}}
}

\author{Chongyu Qu$^1$, Ritchie Zhao$^2$, Ye Yu$^2$, Bin Liu$^2$, Tianyuan Yao$^1$, Junchao Zhu$^1$, Bennett A. Landman$^1$, \\ Yucheng Tang$^2$,  and Yuankai Huo$^1$\thanks{Corresponding author: Yuankai Huo (\href{mailto:yuankai.huo@vanderbilt.edu}{yuankai.huo@vanderbilt.edu})}\\[2mm]
$^1$Vanderbilt University~~~~~~~~$^2$NVIDIA\\ [0.5mm]
}
\maketitle

\begin{abstract}
Quantizing deep neural networks ,reducing the precision (bit-width) of their computations, can remarkably decrease memory usage and accelerate processing, making these models more suitable for large-scale medical imaging applications with limited computational resources. However, many existing methods studied ``fake quantization'', which simulates lower precision operations during inference, but does not actually reduce model size or improve real-world inference speed. Moreover, the potential of deploying real 3D low-bit quantization on modern GPUs is still unexplored. In this study, we introduce a real post-training quantization (PTQ) framework that successfully implements true 8-bit quantization on state-of-the-art (SOTA) 3D medical segmentation models, i.e., U-Net, SegResNet, SwinUNETR, nnU-Net, UNesT, TransUNet, ST-UNet,and VISTA3D. Our approach involves two main steps. First, we use TensorRT to perform fake quantization for both weights and activations with unlabeled calibration dataset. Second, we convert this fake quantization into real quantization via TensorRT engine on real GPUs, resulting in real-world reductions in model size and inference latency. Extensive experiments demonstrate that our framework effectively performs 8-bit quantization on GPUs without sacrificing model performance.  This advancement enables the deployment of efficient deep learning models in medical imaging applications where computational resources are constrained.
The code and models have been released, including U-Net, TransUNet pretrained on the BTCV dataset for abdominal (13-label) segmentation, UNesT pretrained on the Whole Brain Dataset for whole brain (133-label) segmentation, and nnU-Net, SegResNet, SwinUNETR and VISTA3D pretrained on TotalSegmentator V2 for full body (104-label) segmentation.\href{https://github.com/hrlblab/PTQ}{https://github.com/hrlblab/PTQ}.

\end{abstract}

\begin{figure*}[t]
\centerline{\includegraphics[width=0.95\linewidth]{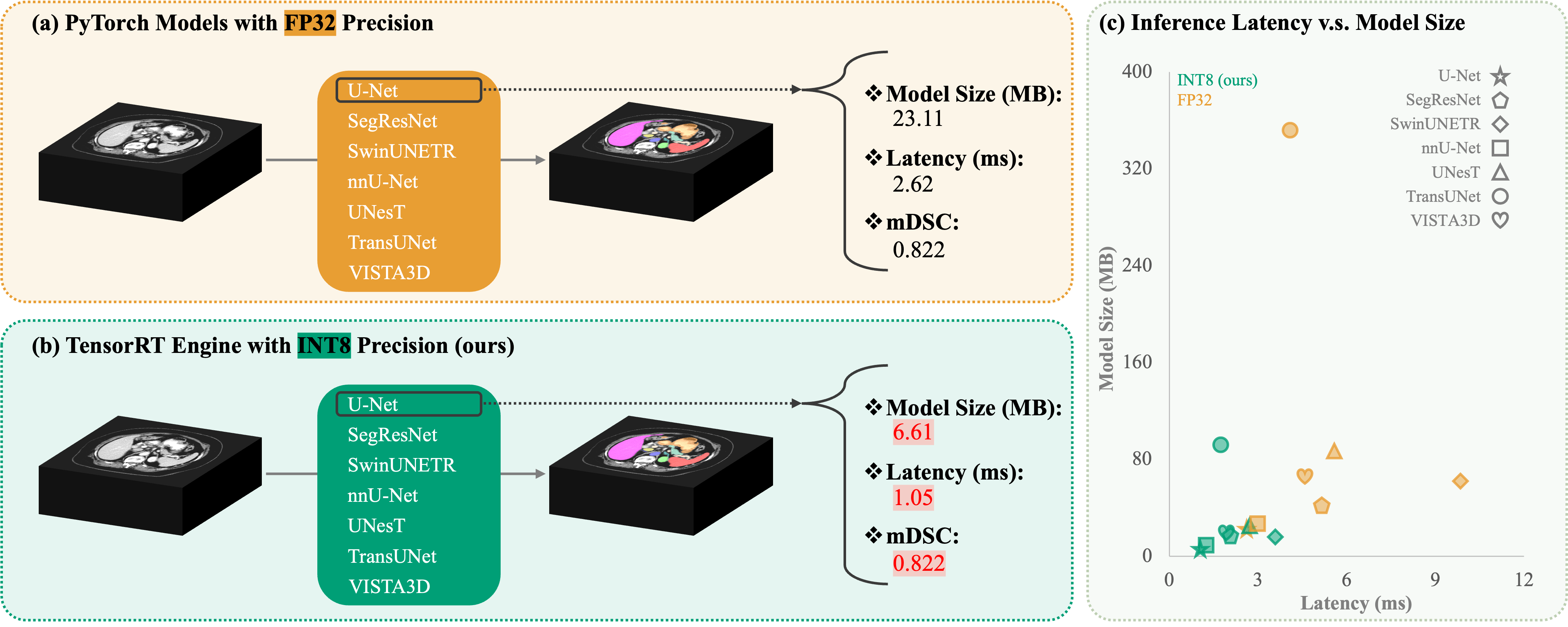}}
\caption{
\textbf{(a) PyTorch Models with FP32 Precision.} Previous 3D medical image segmentation commonly uses FP32 models, which results in larger model sizes, higher computational demands, and slower inference. As medical datasets continue to grow, improving model efficiency becomes increasingly important.
\textbf{(b) TensorRT Engine with INT8 Precision.} We propose a real PTQ framework using NVIDIA TensorRT to convert FP32 models into INT8, enabling notable reductions in both model size and inference latency without compromising performance. For example, U-Net's model size shrinks from 23.11 MB to 6.61 MB, and its inference latency drops from 2.62 ms to 1.05 ms, while maintaining the same mean Dice Score (mDSC) of 0.822.
\textbf{(c) Inference Latency vs. Model Size.} We evaluate seven medical segmentation models, i.e., UNet, SegResNet, SwinUNETR, nnU-Net, UNesT, TransUNet, and VISTA3D, before and after our PTQ framework. Compared with their original FP32 versions (\textcolor{iorange}{orange}), our INT8 models (\textcolor{igreen}{green}) achieve clear smaller model sizes and inference latency, indicating superior efficiency.
}
\label{fig:introduction}
\end{figure*}

\begin{figure*}[t]

\centerline{\includegraphics[width=0.95\linewidth]{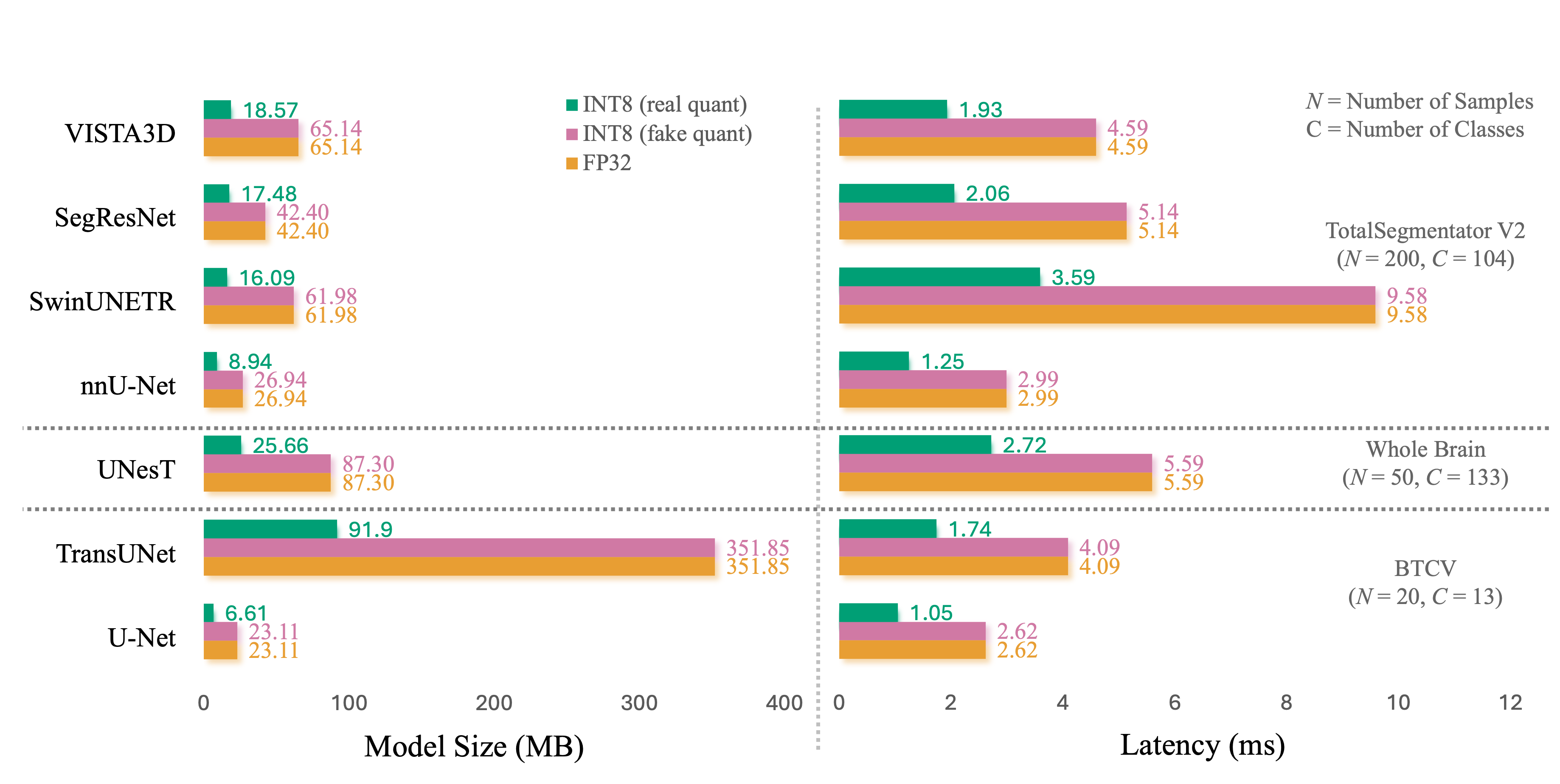}}
\caption{
\textbf{Comparison between real quantization and fake quantization.}
We compare the real quantization and fake quantization on seven medical segmentation models, i.e., VISTA3D, SegResNet, SwinUNETR, nnU-Net, UNesT, TransUNet and U-Net across three datasets with varying sample sizes (\textit{N}) and label counts (\textit{C}), i.e., TotalSegmentator V2 ($N=200, C=104$), Whole Brain ($N=50, C=133$) and BTCV  ($N=20, C=13$). The left panel compares model sizes for INT8 (real quant), INT8 (fake quant), and the original FP32 models, while the right panel compares their inference latencies. As shown, fake quantization only simulates low-precision computation and provides no real-world reduction in model size or latency. In contrast, our real quantization reduces model size by a factor of $2.42\times$ to $3.85\times$ and speeds up inference by $2.05\times$ to $2.66\times$.
}
\label{fig:statistic}
\end{figure*}

\section{Introduction}
\label{sec:introduction}

Deep neural networks have become indispensable in medical imaging tasks, remarkably enhancing diagnostic accuracy and efficiency in tasks such as image classification~\cite{zhang2019medical,kim2022transfer,li2014medical}, segmentation~\cite{patil2013medical,isensee2021nnu,zhou2021towards,chen2018drinet,ji2022amos},  and anomaly detection~\cite{shvetsova2021anomaly,wolleb2022diffusion,fernando2021deep}. Despite their effectiveness, deploying these models in large-scale medical imaging applications poses challenges due to their substantial computational requirements, especially in environments with limited hardware capabilities.

A promising approach to mitigate these challenges is model quantization\cite{polino2018model,zhou2018adaptive,rokh2023comprehensive,gholami2022survey}, which reduces the precision (bit width) of computations within a neural network. By converting high-precision representations (e.g. 32-bit floating point numbers) to lower-precision formats (e.g. 8-bit integers) for both weights and activations, quantization can remarkably decrease memory usage and accelerate processing speeds. This transformation not only makes models more suitable for deployment on devices with constrained resources but also enables faster inference times essential for real-time medical applications.

Quantization methods are broadly categorized into two types, \ding{172} Quantization-Aware Training (QAT) and \ding{173} Post-Training Quantization (PTQ). Quantization-Aware Training (QAT) trains a model by simulating low-precision calculations during both forward and backward passes, allowing the model to adjust to these constraints. As a result, the model can retain its accuracy after being quantized. For example, DeepSeek V3~\cite{liu2024deepseekv3} introduces an FP8 mixed precision training framework and, for the first time, validates its effectiveness on an extremely large-scale model. By leveraging FP8 computation and storage, DeepSeek V3 achieves both accelerated training and reduced GPU memory usage. While QAT can achieve high accuracy at lower precisions, it is time-consuming and requires access to the entire labeled training dataset, which is a notable drawback given the massive size and sensitivity of medical imaging data. PTQ quantizes a pre-trained model without the need for retraining, using only a small set of unlabeled samples to calibrate the network. This makes PTQ more practical for real-world applications, especially when retraining is impractical due to resource constraints or data privacy concerns. Therefore, our focus in this paper is on designing an effective PTQ approach for medical imaging models.

Although previous PTQ methods have achieved notable success in various scenarios, including convolutional neural networks (CNNs)~\cite{wang2024aqa} and vision transformers (ViTs)~\cite{liu2023fgptq,yang2024dopq,tai2024mptq}, a common limitation is their reliance on fake quantization. In this approach, the quantization process is simulated during inference to approximate lower-precision computations, but the underlying model remains in high precision. As a result, there are no actual reductions in model size or meaningful improvements in real-world inference speed. This disconnect between simulated efficiency gains and practical performance limits the benefits of quantization in resource-constrained settings. 

Moreover, recent advancements in deep learning, such as the 1.4B parameter STU-Net~\cite{huang2023stu}, highlight the transformative potential of scaling laws in medical image segmentation, demonstrating that larger models trained on appropriately large datasets achieve superior performance by capturing complex anatomical features. However, the computational and memory demands of such large-scale models present significant challenges for deployment in resource-constrained clinical environments. While PTQ offers a practical solution by reducing precision (e.g., from FP32 to INT8), substantially lowering memory usage and inference latency, existing methods often rely on fake quantization, limiting their effectiveness in many real-world scenarios. Consequently, for models like STU-Net, a real PTQ framework that ensures robustness to precision loss, enabling efficient real-time inference on edge devices without compromising accuracy, remains in high demand. Such an approach would bridge the gap between the theoretical advantages of scaling laws and their practical utility, facilitating scalable and cost-effective deployment of high-capacity models in medical imaging.

In order to fulfill this need, we introduce a real PTQ framework that implements true 8-bit integer (INT8) quantization on SOTA 32-bit floating-point (FP32) medical segmentation models without the need for retraining, as shown in \figureautorefname~\ref{fig:introduction}. Our method involves two key steps. \textbf{Firstly}, we leverage NVIDIA TensorRT\footnote{\href{https://developer.nvidia.com/tensorrt}{NVIDIA TensorRT} is a high-performance deep learning inference optimizer and runtime library that facilitates faster inference on NVIDIA GPUs through graph optimizations, precision calibration, and efficient memory management.} to perform fake quantization of both model weights and activations using unlabeled calibration dataset. \textbf{Secondly}, we convert fake quantized models into real quantized TensorRT engines deployed on actual GPUs. TensorRT applies hardware-specific optimizations that enable efficient low-precision computations, resulting in tangible reductions in model size and faster inference times.

To demonstrate the effectiveness of our real PTQ framework, we compare the model size and inference latency of seven SOTA 3D medical segmentation models, i.e., U-Net~\cite{ronneberger2015u}, TransUNet~\cite{chen2021transunet}, UNesT~\cite{yu2023unest}, nnU-Net~\cite{isensee2021nnu}, SwinUNETR~\cite{hatamizadeh2021swin}, SegResNet~\cite{myronenko20193d}, and VISTA3D~\cite{he2024vista3d}, quantized to INT8 via our framework, with their previously fake quantized INT8 and original FP32 counterparts. These comparisons span three datasets (TotalSegmentator V2~\cite{wasserthal2023totalsegmentator}, Whole Brain~\cite{huo2018spatially}, and BTCV~\cite{landman2015miccai}), as shown in \figureautorefname~\ref{fig:statistic}. 
Unlike fake quantization, which only simulates INT8 computations but still relies on FP32 resources and therefore does not reduce model size or latency, our real PTQ framework yields measurable reduction in model size ($2.42\times$ to $3.85\times$) and inference latency ($2.05\times$ to $2.66\times$), without compromising performance (see \figureautorefname~\ref{fig:introduction}(a) and \figureautorefname~\ref{fig:introduction}(b)).
These results indicate that our PTQ framework is an effective way to reduce resource usage while maintaining high segmentation accuracy in large-scale 3D medical image tasks. In summary, the key contributions of this paper are two-folds:
\begin{itemize}
    \item \textbf{Development of a Practical PTQ framework:}
    We introduce a novel framework for implementing real 3D INT8 PTQ on SOTA medical segmentation models running on modern GPUs, enabling more practical deployment in resource-constrained environments. By offering real reductions in model size, computational demands, and inference latency, our framework significantly enhances efficiency without compromising performance.
    
    \item \textbf{Demonstration of Clinical Applicability:}
    We demonstrate our PTQ framework's robustness by successfully quantizing a broad set of SOTA 3D medical segmentation models, confirming the universal feasibility of real INT8 quantization. As both AI model sizes and dataset sizes continue to grow in clinical practice, our PTQ framework offers a crucial pathway toward resource-efficient, large-scale medical image analysis.
\end{itemize}

\begin{figure*}[t]
\centerline{\includegraphics[width=0.95\linewidth]{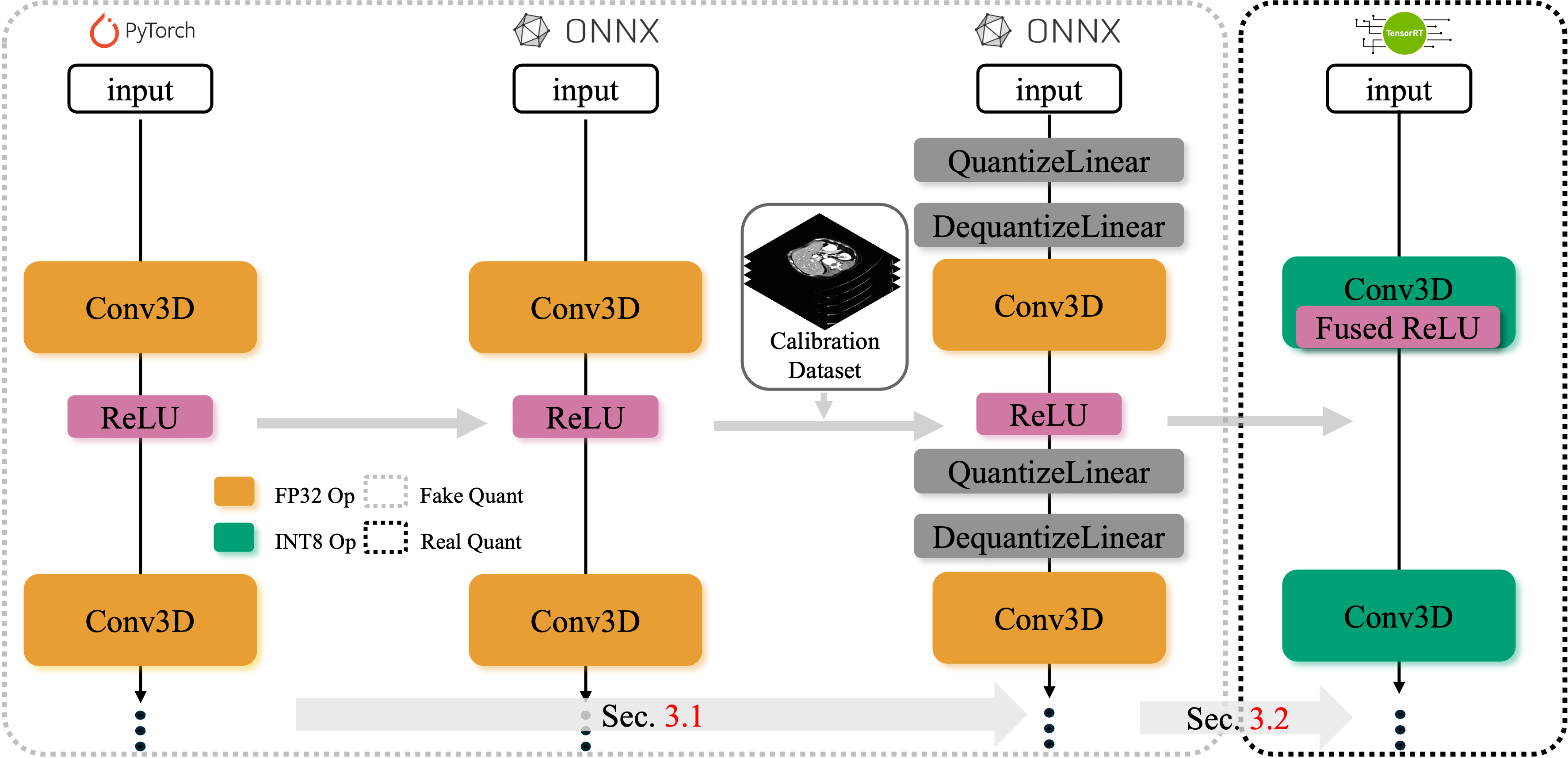}}
\caption{
\textbf{Post-training quantization framework}.  
We first convert the original PyTorch model into the ONNX format. Next, we simulate quantization by adding \texttt{QuantizeLinear} and \texttt{DequantizeLinear} nodes into the ONNX model using a calibration dataset to create a fake quantized model (\S\ref{sec:fake_quant}); this step simulates the INT8 quantization process but still relies on FP32 resources. Finally, we convert this fake quantized model into a real INT8 quantized engine using NVIDIA TensorRT (\S\ref{sec:real_quant}). During this conversion, TensorRT detects the \texttt{QuantizeLinear} and \texttt{DequantizeLinear} nodes to perform actual INT8 quantization, and ReLU layers are fused into preceding layers for performance optimization. }
\label{fig:method}
\end{figure*}

\section{Related Work}
Model quantization has become a critical technique for deploying deep neural networks on resource-constrained hardware by reducing model size and computational demands. Quantization methods are generally categorized into two main approaches: QAT and PTQ.

\smallskip\noindent\textbf{QAT} integrates quantization into the training process, allowing the model to learn and adapt to quantization effects during training. By incorporating discrete constraints directly into the backpropagation algorithm~\cite{lecun1988theoretical}, QAT methods enable networks to maintain high accuracy even at low bit-widths. To facilitate gradient propagation through quantized variables, many QAT methods including XNOR-Net~\cite{rastegari2016xnor}, QIL~\cite{jung2019learning}, 3DQ~\cite{paschali20193dq},  MedQ~\cite{zhang2021medq}, utilize the Straight-Through Estimator (STE)~\cite{yin2019understanding}, which approximates the gradient of the non-differentiable quantization function. While some QAT approaches utilize differentiable approximations of the quantization function during training, which are then replaced with hard quantization during inference. DoReFa-Net ~\cite{zhou2016dorefa} quantizes weights, activations, and even gradients to low bit-widths using differentiable quantizers, DSQ~\cite{gong2019differentiable} introduces a differentiable function that smoothly approximates the quantizer. 
While QAT methods can achieve high accuracy, they have substantial drawbacks. They require intrusion into the training code and additional computational cost due to retraining, which involves large labeled datasets and extensive training time. In the medical imaging domain, this is particularly challenging due to data privacy concerns~\cite{jin2019review,gkoulalas2015introduction,price2019privacy} and the substantial size of medical datasets~\cite{qu2023abdomenatlas,li2024abdomenatlas,li2024medshapenet}.

\smallskip\noindent\textbf{PTQ} method quantizes a pre-trained model without additional training, using only a small set of unlabeled data for calibration. This makes PTQ more practical for real-world applications, especially when retraining is impractical due to resource constraints or data privacy concerns inherent in medical imaging. In the medical imaging domain, two main architectures are prevalent: CNNs and transformers. Numerous successful works have introduced quantization methods specifically designed for these architectures. For CNNs, AdaRound~\cite{nagel2020up} formulates quantization as an optimization problem by introducing a learnable rounding mechanism for weights. BRECQ~\cite{li2021brecq} extends AdaRound by implementing block-wise reconstruction, optimizing quantization parameters across blocks of layers to capture inter-layer dependencies. QDrop~\cite{wei2022qdrop} incorporates a dropout mechanism during the reconstruction process, adding randomness to enhance the robustness and flatness of the optimized model. PD-Quant~\cite{liu2023pd} leverages a prediction difference metric to optimize network blocks globally, introducing global information into the quantization parameter optimization. When quantizing transformers, previous works have focused on solving post-softmax distribution problems and addressing activation distribution outliers in intermediate layers. FQ-ViT~\cite{lin2021fq} simulates the non-uniform distribution of attention maps by employing a log2 quantizer and proposed an integer approximation of the exponential function for softmax quantization, also introducing power-of-two quantization for the layer normalization layer. PTQ4ViT~\cite{yuan2022ptq4vit} introduces twin uniform quantization to handle post-softmax and post-GELU activation distributions. RepQ-ViT~\cite{li2023repq} proposes log$\sqrt{2}$ to better adapt to the distribution of post-softmax results. PTQ4SAM~\cite{lv2024ptq4sam} proposes a bimodal integration strategy, applying a mathematically equivalent sign operation to transform the bimodal distribution of key linear output activations into a more easily quantized normal distribution. Additionally, it introduces adaptive granularity quantization for softmax by searching for the optimal power-of-two base to address substantial variations in post-softmax distributions.
While these previous PTQ methods effectively address the quantization challenges in CNN and transformer architectures, they primarily rely on fake quantization techniques that do not result in actual reductions in model size, computational demand, or inference latency. In this paper, we introduce a real PTQ framework that performs real quantization on modern GPUs, optimizing the inference process and achieving remarkably resource savings.

\section{Methodology}
\label{sec:method}
\noindent\textbf{Overview.} In this section, we present our proposed PTQ framework designed to implement real INT8 quantization on SOTA medical segmentation models on modern GPUs, as shown in Figure~\ref{fig:method}. Our method aims to reduce model size and inference latency without retraining, thus making advanced models more accessible in resource-constrained environments. The framework consists of two main components: first, we leverage NVIDIA TensorRT to perform fake quantization by inserting \texttt{QuantizeLinear} and \texttt{DequantizeLinear} nodes into the Open Neural Network Exchange (ONNX) model with unlabeled calibration dataset (\S\ref{sec:fake_quant}), simulating the quantization process while still relies on FP32 resources. Second, we convert this fake quantized model into a real INT8 quantized TensorRT engine using for efficient deployment on modern GPUs (\S\ref{sec:real_quant}). During this conversion, TensorRT detects the \texttt{QuantizeLinear} and \texttt{DequantizeLinear} nodes to perform real INT8 quantization. 
Our framework addresses the limitations of previous PTQ methods that rely on fake quantization, which simulates lower-precision computations without yielding actual reductions in model size or improvements in inference speed. By converting fake quantization into real quantization, we ensure that the quantized models are not only theoretically efficient but also practically deployable with remarkably resource savings.

\subsection{Perform Fake Quantization on ONNX}
\label{sec:fake_quant}
In the first component of our framework, we perform fake quantization by quantizing both the weights and activations of the pre-trained network with unlabeled calibration dataset. To facilitate this process, we begin by converting the pre-trained PyTorch model into the ONNX format: 
\[
\text{PyTorch Model} \xrightarrow{\text{Export}} \text{ONNX Model}
\]
ONNX is an open standard for representing deep learning models, enabling interoperability between different frameworks and tools. This conversion is essential because it allows us to leverage optimization tools that may not be directly compatible with PyTorch models.

\smallskip\noindent\textbf{Simulation of INT8 Quantization with ModelOpt.} After converting the model to ONNX, we use the NVIDIA TensorRT Model Optimizer (ModelOpt) to simulate the INT8 quantization process:
\[
\text{ONNX Model} \xrightarrow{\text{ModelOpt}} \text{Fake Quantized Model}
\]
The ModelOpt analyzes the computational graph of the model and inserts \texttt{QuantizeLinear} and \texttt{DequantizeLinear} nodes where appropriate. The \texttt{QuantizeLinear} nodes simulate the conversion of FP32 weights and activations into INT8 values by applying scale factors and zero-points determined during calibration, while the \texttt{DequantizeLinear} nodes convert these simulated INT8 values back into FP32 for subsequent computations. By inserting these nodes, we prepare the model for real quantization in the next step, where NVIDIA TensorRT can detect these nodes and convert the operations into real INT8 computations on modern GPUs. The calibration process involves mapping floating-point values to discrete integer levels that can be represented with lower bit-width precision. For a given floating-point value $x$ the quantized value $x_q$ can be defined as:
\begin{equation}
    x_q = \text{clamp}\left(\left\lfloor \frac{x}{s} \right\rceil+z,0,2^k-1\right),
\end{equation}
where the scale factor $s$ and zero-point $z$ are quantized parameters determined during calibration by
\begin{equation}
    s = \frac{x_{max}-x_{min}}{2^k-1},
\end{equation}
\begin{equation}
    z = - \left\lfloor \frac{x_{min}}{s} \right\rceil,
\end{equation}
where $k$ is the number of quantization bits, $\left\lfloor \cdot \right\rceil$ denotes rounding to the nearest integer, and $\text{clamp}(\cdot,a,b)$ restrict the value within the range $[a,b]$. Here, $x_{min}$ and $x_{max}$ represent the minimum and maximum values of $x$ observed in the calibration dataset.
\subsection{Converting to Real Quantization Using NVIDIA TensorRT}
\label{sec:real_quant}
In the second step of our framework, we transform the fake quantized ONNX model into a real INT8 quantized engine optimized for efficient deployment using NVIDIA TensorRT. This involves converting the \texttt{QuantizeLinear} and \texttt{DequantizeLinear} nodes inserted during the simulation phase into actual INT8 operations, enabling true real computations on modern GPUs. 

\smallskip\noindent\textbf{Building the TensorRT Engine.} We input the fake quantized ONNX model into TensorRT, which automatically detects the \texttt{QuantizeLinear} and \texttt{DequantizeLinear} nodes, and replaces the corresponding operations with optimized INT8 kernels:
\[
\text{Fake Quantized Model} \xrightarrow{\text{TensorRT}} \text{INT8 Engine}
\]
By selecting the most efficient execution kernels optimized for INT8 precision, TensorRT fully leverages the capabilities of NVIDIA GPUs.
By converting fake quantization into real quantization, we achieve noticeable reductions in model size and considerable improvements in inference speed. The resulting TensorRT engine executes supported operations in true INT8 precision, providing a practical solution for deploying advanced deep learning models in resource-constrained environments.
\begin{table*}[t]
\caption{\textbf{Quantization results of SOTA medical segmentation models.} We evaluate our PTQ framework using seven SOTA medical segmentation models, i.e. U-Net, TransUNet, UNesT, nnU-Net, SwinUNETR, SegResNet and VISTA3D, across three dataset with different number of samples (\textit{N}) and number of classes (\textit{C}) i.e., BTCV (\textit{N} = 20, \textit{C} = 13), Whole Brain Segmentation (\textit{N} = 50, \textit{C} =133) and TotalSegmentator V2 (\textit{N} = 200, \textit{C} = 104). All models are converted to TensorRT engines for both FP32 and INT8, and performance is measured by the mean Dice Similarity Coefficient (mDSC). Compared with their FP32 counterparts, our PTQ framework reduces model size by factors of $2.42\times$ (42.42/17.48) to $3.85\times$ (61.98/16.09), and inference latency by $2.05\times$ (5.59/2.72) to $2.66\times$ (9.58/3.59). These improvements come without sacrificing performance, as the mDSC remains nearly unchanged after quantization.
}
\centering
\scriptsize
\begin{tabular}{
    P{0.15\linewidth}               
    p{0.14\linewidth}               
    P{0.085\linewidth}P{0.085\linewidth}|  
    P{0.085\linewidth}P{0.085\linewidth}|  
    P{0.085\linewidth}P{0.085\linewidth}   
}
\toprule
\multirow{2}{*}{Dataset} & 
\multirow{2}{*}{AI Models} & 
\multicolumn{2}{c|}{Model Size (MB)} &
\multicolumn{2}{c|}{Latency (ms)} &
\multicolumn{2}{c}{mDSC} \\
\cmidrule{3-8}
& & FP32 & INT8 & FP32 & INT8 & FP32 & INT8 \\
\midrule

\multirow{2}{*}{
    \parbox{2.3cm}{\centering 
        BTCV~\cite{landman2015miccai}\\
        (\textit{N} = 20, \textit{C} = 13)
    }
}
 & U-Net~\cite{ronneberger2015u}    
   & 23.11  & \cellcolor{igray!5}6.61   
   & 2.62    & \cellcolor{igray!5}1.05  
   & 0.822   & \cellcolor{igray!5}0.822 \\
 & TransUNet~\cite{chen2021transunet}
   & 351.85 & \cellcolor{igray!5}91.90  
   & 4.09    & \cellcolor{igray!5}1.74  
   & 0.816   & \cellcolor{igray!5}0.816 \\
\midrule

    \parbox{2.5cm}{\centering 
       Whole Brain~\cite{huo2018spatially}\\
       (\textit{N} = 50, \textit{C} = 133)
    }
 & UNesT~\cite{yu2023unest}     
   & 87.30  & \cellcolor{igray!5}25.66  
   & 5.59    & \cellcolor{igray!5}2.72 
   & 0.702   & \cellcolor{igray!5}0.701 \\
\midrule

\multirow{4}{*}{
    \parbox{2.5cm}{\centering 
       TotalSeg V2~\cite{wasserthal2023totalsegmentator}\\
       (\textit{N} = 200, \textit{C} = 104)
    }
}
 & nnU-Net~\cite{isensee2021nnu}     
   & 26.94  & \cellcolor{igray!5}8.94  
   & 2.99    & \cellcolor{igray!5}1.25 
   & 0.901   & \cellcolor{igray!5}0.895 \\
 & SwinUNERT~\cite{hatamizadeh2021swin}     
   & 61.98  & \cellcolor{igray!5}16.09  
   & 9.85    & \cellcolor{igray!5}3.59 
   & 0.878   & \cellcolor{igray!5}0.877 \\
 & SegResNet~\cite{myronenko20193d}     
   & 42.40  & \cellcolor{igray!5}17.48 
   & 5.14    & \cellcolor{igray!5}2.06 
   & 0.882   & \cellcolor{igray!5}0.879 \\
 & VISTA3D~\cite{he2024vista3d}     
   & 65.14  & \cellcolor{igray!5}18.57  
   & 4.59    & \cellcolor{igray!5}1.93 
   & 0.893   & \cellcolor{igray!5}0.891 \\
\bottomrule
\end{tabular}
\begin{tablenotes}
    \item \textit{N} = Number of Samples \quad\quad
    \textit{C} = Number of Classes
\end{tablenotes}
\label{tab:segmentation}
\end{table*}

\section{Experiments \& Results}
\label{sec:result}
\subsection{Dataset.}
\label{sec:dataset}
\smallskip\noindent\textbf{BTCV}\cite{landman2015miccai} consists of 70 CT volumes with 13 labeled anatomies. They are randomly selected from a combination of an ongoing colorectal cancer chemotherapy trial, and a retrospective ventral hernia study. Of these, 50 CT volumes, which are publicly available through the MICCAI 2015 Multi-Atlas Labeling Challenge, are used to pre-train our U-Net and TransUnet models. The remaining 20 CT volumes are used for evaluation.

\smallskip\noindent\textbf{TotalSegmentator V2}\cite{wasserthal2023totalsegmentator} includes 1,228 full-body CT volumes with 117 labeled anatomies, created by the Department of Research and Analysis at University Hospital Basel. We use 200 of these CT volumes to evaluate nnU-Net, SegResNet, SwinUNETR, and VISTA3D across 104 labels. The remaining 1,028 volumes are used to pre-train these models.

\smallskip\noindent\textbf{Whole Brain Segmentation Dataset}\cite{huo2018spatially} combines 4,859 T1-weighted (T1w) MRI volumes collected from eight different sites, with segmentation labels generated by a multi-atlas segmentation pipeline. Among these volumes, 50 come from the Open Access Series on Imaging Studies (OASIS) dataset~\cite{marcus2007open} and have been manually traced to 133 labels based on the BrainCOLOR protocol~\cite{klein2010open} by Neuromorphometrics Inc. We use these 50 manually-labeled scans to evaluate our UNesT models, while the remaining 4,809 scans are used for pre-training.

\subsection{Implementation Details.}
\label{sec:implementation}
U-Net and TransUNet are pre-trained and evaluated on the BTCV dataset, UNesT is pre-trained and evaluated on the Whole Brain Segmentation dataset, nnU-Net, SegResNet, SwinUNETR, and VISTA3D are pre-trained and evaluated on TotalSegmentator V2. Except for nnU-Net, which follows its default training plan and original learning rate, these models share the same data augmentation and pre-processing steps, and are trained on a single NVIDIA RTX 4090 GPU with an input volume size of $96\times96\times96$. They employ the Adam optimizer starting at 1$e$-4 and a weight decay of 1$e$-5, , with the learning rate dynamically adjusted based on the combined Dice and Cross Entropy (DiceCE) Loss. After training, all models are quantized into INT8 engines using NVIDIA TensorRT, retaining FP32 for input and output, and both quantization and evaluation perform on a single NVIDIA RTX 4090 GPU.
Segmentation accuracy is measured via the mean Dice Similarity Coefficient (mDSC), and to compare model efficiency between INT8 and FP32 versions, we monitor model size, GPU memory usage during inference, and inference latency.

\begin{table}[t]
\caption{\textbf{GPU Memory Usage for U-Net and TransUNet on BTCV.} We compare the GPU memory usage for U-Net and TransUNet across PyTorch FP32, TensorRT FP32, and TensorRT INT8 (via our PTQ framework). The results show marked memory savings when leveraging TensorRT optimizations, with further reductions through our real INT8 PTQ, thereby saving computational resource and improving overall model efficiency. 
}

\centering
\scriptsize
\begin{tabular}{
    P{0.18\linewidth}               
    p{0.14\linewidth}               
    P{0.14\linewidth}P{0.14\linewidth}P{0.14\linewidth}
}
\toprule
\multirow{2}{*}{Dataset} & 
\multirow{2}{*}{AI Models} & 
\multicolumn{3}{c}{GPU Memory Usage (MB)} 
\\
\cmidrule{3-5}
& & FP32 (PyTorch) & FP32 (TensorRT) & INT8 (TensorRT)\\
\midrule

\multirow{2}{*}{
    \parbox{1.7cm}{\centering 
        BTCV~\cite{landman2015miccai}\\
        (\textit{N} = 20, \textit{C} = 13)
    }
}
 & U-Net~\cite{ronneberger2015u}     
   & 6391.25  & 1893.28  
   & 1787.28  \\
& TransUNet~\cite{chen2021transunet}     
   & 6331.25  & 2235.28  
   & 1873.28  \\
\bottomrule
\end{tabular}
\begin{tablenotes}
    \item \textit{N} = Number of Samples \quad\quad
    \textit{C} = Number of Classes
\end{tablenotes}
\label{tab:gpu}
\end{table}

\subsection{Quantization Results of Segmentation Models}
\label{sec:segmentation}
We evaluate our PTQ framework using seven SOTA medical segmentation models, i.e. U-Net, TransUNet, UNesT, nnU-Net, SwinUNETR, SegResNet and VISTA3D, across three dataset with different number of samples (\textit{N}) and number of classes (\textit{C}) i.e., BTCV (\textit{N} = 20, \textit{C} = 13), Whole Brain Segmentation (\textit{N} = 50, \textit{C} =133) and TotalSegmentator V2 (\textit{N} = 200, \textit{C} = 104). To eliminate inconsistencies between libraries (PyTorch vs. TensorRT), all models are converted to TensorRT engines for both FP32 and INT8. As shown in \tableautorefname~\ref{tab:segmentation}, the INT8 quantized models achieve $2.42\times$ (42.42/17.48) to $3.85\times$ (61.98/16.09) reductions in model size and $2.05\times$ (5.59/2.72) to $2.66\times$ (9.58/3.59) reductions in inference latency, while maintaining the same mDSC performance as their FP32 counterparts. These results demonstrate that our PTQ framework delivers real-world gains in efficiency, especially beneficial when working with large-scale datasets that typically require a large amount of inference time.

We also measure GPU memory usage for U-Net and TransUNet on BTCV, as shown in \tableautorefname~\ref{tab:gpu}.  Our PTQ framework uses NVIDIA TensorRT, which automatically optimizes models to conserve computational resources during inference. Compared with FP32 models on PyTorch, the FP32 TensorRT engines reduce GPU memory usage by $3.37\times$ (6391.25/1893.28) and $2.83\times$ (6331.25/2235.28) for U-Net and TransUNet, respectively. After quantizing to INT8 with our PTQ framework, these reductions increase to $3.57\times$ (6391.25/1787.28) and $3.37\times$ (6331.25/1873.28). These results show that our PTQ framework leverages TensorRT’s optimizations to achieve real-world computational savings and improve model efficiency.

\subsection{Scaling Analysis}
\label{sec:scaling}
To evaluate the impact of scaling effectiveness on TotalSegmentator V2 (\textit{N} = 200, \textit{C} = 104), we compare the performance and efficiency of STU-Net-S (14M parameters) and STU-Net-H (1.4B parameters)~\cite{huang2023stu}. As shown in \tableautorefname~\ref{tab:scaling}, by scaling the model size, segmentation accuracy (measured by mDSC) increases from 0.837 to 0.869,  demonstrating the benefits of scaling for capturing complex anatomical structures. 

Despite the performance gains, scaling introduces higher computational demands. On TotalSegmentator V2, inference latency for STU-Net increases from 2.59 ms to 98.45 ms, making it prohibitively slow when processing large-scale datasets. After applying our PTQ framework, the INT8 quantized STU-Net-H model size decreases by $3.65\times$ (1457.33/398.41), and its inference latency drops by $3.26\times$ (98.45/30.15), while maintaining comparable mDSC scores to the FP32 counterpart. Compared with smaller STU-Net-S that already achieve a latency of under 10 ms, our PTQ approach yields even more pronounced gains for large-scale models like STU-Net-H, cutting latency by an additional 68.3 ms. Furthermore, the INT8 quantized STU-Net-H achieves a higher compression ratio in terms of model size ($3.65\times$) than STU-Net-S ($2.71\times$), because smaller models are more impacted by overheads and non-quantizable layers, whereas larger models have a greater proportion of parameters that benefit from quantization. Consequently, our PTQ framework proves particularly beneficial for large-scale models, making it a powerful solution for deploying resource-efficient deep learning systems.


These results demonstrate that our PTQ framework effectively addresses the computational challenges of scaling, enabling the deployment of large-scale models like STU-Net in resource-constrained environments. This highlights the synergy between scaling laws and quantization for advancing medical image segmentation.
\begin{table*}[t]
\caption{\textbf{Quantization results of STU-Net on TotalSegmentator V2.} We evaluate the impact of scaling effectiveness using STU-Net (scaling up from 14M to 1.4B parameters) on TotalSegmentator V2 ($N$ = 200, $C$ = 104). The segmentation performance (measured by mDSC) increases from 0.837 to 0.869, demonstrating the benefits of scaling for capturing complex anatomical structures. Then, by applying our PTQ framework, the INT8 quantized STU-Net-H shrinks by $3.65\times$ (1457.33/398.41) and reduces inference latency by $3.26\times$ (98.45/30.15), while preserving competitive segmentation performance as its FP32 counterpart. The results indicate that our PTQ framework effectively addresses the computational challenges of scaling.
}
\centering
\scriptsize



\begin{tabular}{
    P{0.15\linewidth}               
    p{0.14\linewidth}               
    P{0.085\linewidth}P{0.085\linewidth}|  
    P{0.085\linewidth}P{0.085\linewidth}|  
    P{0.085\linewidth}P{0.085\linewidth}   
}
\toprule
\multirow{2}{*}{Dataset} & 
\multirow{2}{*}{AI Models} & 
\multicolumn{2}{c|}{Model Size (MB)} &
\multicolumn{2}{c|}{Latency (ms)} &
\multicolumn{2}{c}{mDSC} \\
\cmidrule{3-8}
& & FP32 & INT8 & FP32 & INT8 & FP32 & INT8 \\
\midrule

\multirow{2}{*}{
    \parbox{2.5cm}{\centering 
       TotalSeg V2~\cite{wasserthal2023totalsegmentator}\\
       (\textit{N} = 200, \textit{C} = 104)
    }
}
 & STU-Net-S~\cite{huang2023stu}    
   & 14.60  & \cellcolor{igray!5}5.38   
   & 2.59    & \cellcolor{igray!5} 1.02
   & 0.837   & \cellcolor{igray!5}0.835 \\
 & STU-Net-H
   & 1457.33 & \cellcolor{igray!5}398.41  
   & 98.45    & \cellcolor{igray!5}30.15  
   & 0.869   & \cellcolor{igray!5}0.866 \\
\bottomrule
\end{tabular}
\begin{tablenotes}
    \item \textit{N} = Number of Samples \quad\quad
    \textit{C} = Number of Classes
\end{tablenotes}
\label{tab:scaling}
\end{table*}

\section{Discussion}
\label{sec:discussion}
The results of our experiments demonstrate that our proposed PTQ framework effectively reduces model size, computational demands and inference latency without compromising model performance. Notably, the quantized INT8 models maintain segmentation accuracy comparable to their FP32 counterparts, as indicated by the nearly unchanged mDSC. This finding is noteworthy because it challenges the common concern that quantization, particularly at low bit-widths like INT8, inherently leads to a degradation in model performance.

\smallskip\noindent\textbf{Robustness Across Models and Datasets.}
The effectiveness of our framework across SOTA medical segmentation models, i.e., U-Net, TransUNet, UNesT, nnU-Net, SwinUNETR, SegResNet and VISTA3D, and datasets, i.e., BTCV, Whole Brain Segmentation dataset and TotalSegmentator V2, underscores its generalizability. These models vary in architecture and complexity, and the datasets cover a range of anatomical structures and imaging modalities. The consistent performance across these variations indicates that our PTQ framework is robust and adaptable to various AI models used in medical imaging domain.

\smallskip\noindent\textbf{Clinical Applications.}
Diagnostic accuracy is critically important in the medical domain, as it directly affects patient care and treatment outcomes. Maintaining model performance after quantization is essential to ensure that efficiency gains do not compromise clinical effectiveness. Our successful application of INT8 quantization to complex segmentation tasks involving high-resolution medical images demonstrates that our framework preserves diagnostic accuracy and can be safely integrated into clinical workflows. This integration has the potential to improve patient outcomes through faster and more efficient diagnostics.

\smallskip\noindent\textbf{Potential Limitations and Future Work.}
Despite the promising results, there are certain limitations to our PTQ framework that need to be addressed. One notable limitation stems from the use of NVIDIA TensorRT for converting fake quantization into real INT8 computations. TensorRT may not fully support models with dynamic blocks or layers that require runtime flexibility, such as those involving variable input sizes or conditional operations. These dynamic architectures can pose compatibility issues with TensorRT's optimization and quantization processes, potentially limiting the applicability of our framework to a subset of AI models. Addressing this limitation would involve enhancing the compatibility of TensorRT with dynamic model components or exploring alternative optimization tools that can handle such architectures effectively.

For future work, a promising direction is to focus on quantization to even lower bit-widths, such as 4-bit integer (INT4). Exploring INT4 quantization has the potential to further reduce model size and computational requirements, offering additional benefits for deployment in extremely resource-constrained environments. However, achieving INT4 quantization without compromising model accuracy presents substantial challenges. The reduced precision can introduce considerable quantization errors, leading to performance degradation. Developing advanced quantization techniques, such as adaptive quantization strategies or error compensation methods, will be crucial to maintain model performance at these lower precisions. Additionally, researching hardware accelerators optimized for INT4 computations could enhance the practical feasibility of deploying such quantized models in real-world medical imaging applications.
\section{Conclusion}
\label{sec:conclusion}
We have introduced a PTQ framework that achieves real INT8 quantization for SOTA 3D AI models in medical imaging applications. Our framework effectively reduces real-world model size, computational requirements and inference latency without compromising segmentation accuracy on modern GPU, as evidenced by the maintained mDSC comparable to full-precision models. The framework's robustness across diverse set of AI architectures, ranging from CNN based to transformer based models, and a wide variety of medical imaging datasets. These datasets are collected from multiple hospitals with distinct imaging protocols, cover different body regions (such as brain, abdomen, or full body), and include multiple imaging modalities (CT and MRI). collectively, these factors highlight our PTQ framework’s strong generalizability and adaptability for a broad spectrum of medical imaging tasks.

By preserving diagnostic accuracy while enhancing computational efficiency, our PTQ framework holds considerable potential for clinical integration. It enables the deployment of advanced AI models in resource-constrained environments, facilitating faster and more efficient diagnostics without compromising patient care.

\smallskip\noindent\textbf{Acknowledgments.} This research was supported by NIH R01DK135597 (Huo), DoD HT9425-23-1-0003 (HCY), and KPMP Glue Grant. This work was also supported by Vanderbilt Seed Success Grant, Vanderbilt Discovery Grant, and VISE Seed Grant. This project was supported by The Leona M. and Harry B. Helmsley Charitable Trust grant G-1903-03793 and G-2103-05128. This research was also supported by NIH grants R01EB033385, R01DK132338, REB017230, R01MH125931, and NSF 2040462. We extend gratitude to NVIDIA for their support by means of the NVIDIA hardware grant. This work was also supported by NSF NAIRR Pilot Award NAIRR240055.

{\small
\bibliographystyle{ieee_fullname}
\bibliography{refs,zzhou}
}

\end{document}